# A Component Based Heuristic Search Method with Evolutionary Eliminations for Hospital Personnel Scheduling

Jingpeng Li, Uwe Aickelin and Edmund K. Burke

School of Computer Science, The University of Nottingham, Nottingham, NG8 1BB, United Kingdom

{jpl, uxa, ekb}@cs.nott.ac.uk

Nurse rostering is a complex scheduling problem that affects hospital personnel on a daily basis all over the world. This paper presents a new component-based approach with evolutionary eliminations, for a nurse scheduling problem arising at a major UK hospital. The main idea behind this technique is to decompose a schedule into its components (i.e. the allocated shift pattern of each nurse), and then to implement two evolutionary elimination strategies mimicking natural selection and natural mutation process on these components respectively to iteratively deliver better schedules. The worthiness of all components in the schedule has to be continuously demonstrated in order for them to remain there. This demonstration employs an evaluation function which evaluates how well each component contributes towards the final objective. Two elimination steps are then applied: the first elimination eliminates a number of components that are deemed not worthy to stay in the current schedule; the second elimination may also throw out, with a low level of probability, some worthy components. The eliminated components are replenished with new ones using a set of constructive heuristics using local optimality criteria. Computational results using 52 data instances demonstrate the applicability of the proposed approach in solving real-world problems.

## 1. Introduction

Employee scheduling has been widely studied for more than 40 years. The following survey papers give an overview of the area: Bradley and Martin 1990; Ernst et al. 2004a and 2004b. Employee scheduling can be thought of as the problem of assigning employees to shifts or duties over a scheduling period so that certain organizational and personal constraints are satisfied. It involves the construction of a schedule for each employee within an organization in order for a set of tasks to be fulfilled. In the domain of healthcare, this is particularly challenging because of the presence of a range of different staff requirements on different days and shifts. Unlike many other organizations, healthcare institutions work twenty-four hours a day for every single day of the year. Irregular shift work has an effect on the nurses' well being and job satisfaction (Mueller and McCloskey 1990). The extent to which the staff roster satisfies the staff can impact significantly upon the working environment.

Automatic approaches have significant benefits in saving administrative staff time and also generally improve the quality of the schedules produced. However, until recently, most personnel scheduling problems in hospitals were solved manually (Silvestro and Silvestro 2000). Scheduling by hand is usually a very time consuming task. Without an automatic tool to generate schedules and to test the quality of a constructed schedule, planners often have to use very straightforward constraints on working time and idle time in the recurring process. Even when hospitals have computerized systems, testing and graphical features are often used but automatic schedule generation features are still not common. Moreover, there is a growing realisation that the automated generation of personnel schedules within healthcare can provide significant benefits and savings. In this paper, we focus on the development of new techniques for automatic nurse rostering systems. A general overview of various approaches for nurse rostering can be found in Sitompul and Randhawa (1990), Cheang et al. (2003) and Burke et al. (2004).

Most real world nurse rostering problems are extremely complex and difficult. Tien and Kamiyama (1982), for example, say nurse rostering is more complex than the travelling salesman problem due to the additional constraint of total number of working days within the scheduling period. Since the 1960's, many papers have been published on various aspects of nurse rostering. Early papers (Warner and Prawda 1972, Miller et al. 1976) attempted to solve the problem by using mathematical programming models. However, computational difficulties exist with these approaches due to the enormous size of the search space. In addition, for most real problems, the goal of finding the 'optimal' solution is not only completely infeasible, but also largely meaningless. Hospital administrators normally want to quickly create a high quality schedule that satisfies all hard constraints and as many soft constraints as possible.

The above observations have led to a number of other attempts to solve real world nurse rostering problems. Several heuristic methods have been developed (e.g., Blau 1985, Anzai and Miura 1987). In the 1980's and later, artificial intelligence methods for nurse rostering, such as constraint programming (Meyer auf'm Hofe 2001), expert systems (Chen and Yeung 1993) and knowledge based systems (Beddoe and Petrovic 2006) were investigated with some success. In the 1990's and later, many of the papers tackle the problem with meta-heuristic methods, which include simulated annealing (Brusco and Jacobs 1995), variable neighbourhood search (Burke et al., 2004), tabu search (Dowsland 1998, Burke et al. 1999) and evolutionary methods (Burke et al. 2001, Kawanaka et al. 2001). In very recent years, there have been increasing interests in the study of mathematical programming based heuristics (Bard and Purnomo 2006 and 2007, Beliën and Demeulemeester 2006) and the study of hyper-heuristics (Burke et al. 2003, Ross 2005) for the problem (Burke et al. 2003, Özcan 2005).

This paper tackles a nurse rostering problem arising at a major UK hospital (Aickelin and Dowsland 2000; Dowsland and Thompson 2000). Its target is to create weekly schedules for wards of nurses by assigning each nurse one of a number of predefined shift patterns in the most efficient way. Besides the traditional approach of Integer Linear Programming (Dowsland and Thompson 2000), a number of meta-heuristic approaches have been explored for this problem. For example, in (Aickelin and Dowsland 2000 and 2003, Aickelin and White 2004) various approaches based on genetic algorithms are presented. In (Li and Aickelin 2004) an approach based on a learning classifier system is investigated. In (Burke et al. 2003) a tabu search hyperheuristic is introduced, and in (Aickelin and Li 2007) an estimation of distribution algorithm is described. In this paper we will report a new component-based heuristic search approach with evolutionary eliminations, which implements optimization on the components within single schedules. This approach combines the features of iterative improvement and constructive perturbation with the ability to avoid getting stuck at local minima. Similar to the ruin and recreate principle reported in (Schrimpf et al. 2000), our approach applies a new method to destroy a part of a solution and then reconstruct it.

The framework of our new algorithm is an iterative improvement heuristic, in which the steps of Evaluation, Elimination-I, Elimination-II and Reconstruction are executed in a loop until a stopping condition is reached. In the Evaluation step, a current complete schedule is first decomposed into assignments for individual nurses, and then the assignment for each nurse is evaluated by a function based upon both hard constraints and soft constraints. In the Elimination-I step, some nurses are marked as 'rescheduled' and their assignments are removed from the schedule according to the evaluating values of their assignments. In the Elimination-II step, each remaining nurse still has a small chance to be rescheduled, disregarding the evaluating value of his/her assignment. Finally, in the Reconstruction step, a refined greedy heuristic is designed to repair a partial (candidate) solution and the obtained complete solution is fed into the Evaluation step again to repeat the loop.

Our proposed approach belongs to the general class of local search methods. In particular, it is somewhat similar to the Iterated Local Search algorithm (Lourenco et al. 2002): they include a solution elimination phase and an improvement phase. However, they differ in the way in which these two phases are implemented: The purpose of elimination in Iterated Local Search is to transform one complete solution into another complete solution. This serves as the starting point for the local heuristics which follow. However, the aim of the elimination in our method is to transform one complete solution into a partial solution which is then fed into the reconstruction heuristics for repair.

The rest of this paper is organized as follows. Section 2 gives an overview of the nurse rostering problem, and introduces the general framework of our methodology. Section 3 presents our algorithm for nurse rostering. Benchmark results using real-world data sets collected from a major UK hospital are presented in section 4. Concluding remarks are in section 5.

# 2  Preliminaries

## 2.1  The Nurse Rostering Problem

The nurse rostering problem tackled in this paper is concerned with creating weekly schedules for wards of up to 30 nurses at a large UK hospital. These schedules have to meet the demand for a minimum number of nurses of different grades on each shift, whilst being seen to be fair by the staff concerned and satisfying working contracts. The fairness objective is achieved by meeting as many of the nurses' requests as possible and considering historical information (e.g. previous weekends) to ensure that unsatisfied requests and unpopular shifts are evenly distributed. In our model, the day is partitioned into three shifts: two types of day shift known as 'earlies' and 'lates', and a longer night shift. Due to hospital policy, a nurse would normally work either days or nights in a given week (but not both), and because of the difference in shift length, a full week's work would normally include more days than nights. However, some special nurses work other mixtures and the problem can hence not simply be decomposed into days and nights.

However, as described in Dowsland and Thompson (2000), the problem can be split into three independent stages. The first uses a knapsack model to ensure that there are sufficient nurses to meet the covering constraints. If not, additional nurses (agency staff) are allocated to the ward, so that the problem tackled in the second phase is always feasible. The second stage is the most difficult and involves allocating the actual days or nights a nurse works. Once this has been decided, a third phase uses a network flow model (Ahuja et al. 1993) to allocate those on days to 'earlies' and 'lates'. Since stages 1 and 3 can be solved quickly, this paper is only concerned with the highly constrained second step.

The days or nights that a nurse could work in one week define the set of feasible weekly work patterns (i.e. shift patterns) for that nurse. Each shift pattern can be represented as a 0-1 vector with 14 elements, where the first 7 elements represent the 7 days of the week and the last 7 elements the corresponding 7 nights of the week. A '1' or '0' in the vector denotes a scheduled day/night "worked" or "not worked". For example, (1111100 0000000) would be a pattern where the nurse works the first 5 days and no nights. In total, the hospital allows just under 500 such shift patterns. A specific nurse's contract usually allows 50 to 100 of these. Depending on the nurses' preferences, the recent history of patterns worked, and the overall attractiveness of the pattern, a preference cost is allocated to each nurse-shift pattern pair. These values were set in close consultation with the hospital and range from 0 (perfect) to 100 (unacceptable), with a bias to lower values. Due to the introduction of these preference costs which takes into account historic information (e.g. weekends worked in previous weeks), we are able to reduce the planning horizon from the original five weeks to the current one week without affecting solution quality. Further details about the problem can be found in Dowsland (1998).

The problem can be formulated as follows.

Decision variables:

$x_{ij}$ =1 if nurse i works shift pattern j, 0 otherwise.

Parameters:

m = Number of possible shift patterns;

n = Number of nurses;

g = Number of grades;

$a_{jk}$ = 1 if shift pattern j covers period k, 0 otherwise;

$q_{is}$ = 1 if nurse i is of grade s or higher, 0 otherwise;

$p_{ij}$ = Preference cost of nurse i working shift pattern j;

$R_{ks}$ = Demand for nurses with grade s on period k;

A(i) = Set of feasible shift patterns for nurse i.

Target function:

$$\text{Min} \sum_{i=1}^{n} \sum_{j \in A(i)} p_{ij} x_{ij} \; . \tag{1}$$

Subject to:

$$\sum_{j \in A(i)} x_{ij} = 1, \forall i \in \{1,...,n\} \; , \tag{2}$$

$$\sum_{j \in A(i)} \sum_{i=1}^{n} q_{is} a_{jk} x_{ij} \ge R_{ks}, \forall k \in \{1,...,14\}, s \in \{1,...,g\} \; . \tag{3}$$

The constraints outlined in (2) ensure that every nurse works exactly one shift pattern from his/her feasible set. The constraints represented by (3) ensure that the demand for nurses is fulfilled for every grade on every day and night and in line with hospital policy more nurses than necessary may work during any given period. In practice, there is an acute shortage of nurses and actual overstaffing is very rare. Note that the definition of $q_{is}$ allows that higher graded nurses can substitute those at lower grades if necessary. This problem can be regarded as a multiple-choice set-covering problem. The sets are given by the shift pattern vectors and the objective is to minimize the cost of the sets needed to provide sufficient cover for each shift at each grade. The constraints described in (2) enforce the choice of exactly one pattern (set) from the alternatives available for each nurse.

## 2.2   General Description of the Component Based Heuristic Method with Evolutionary Elimination (CHEE)

The basic methodology iteratively operates the steps of Evaluation, Elimination-I, Elimination-II and Reconstruction in a loop on one solution (see the pseudo code presented in Figure 1). At the beginning of the loop, an Initialization step is used to obtain a starting solution and initialize some input parameters (e.g. stopping conditions). In the Evaluation step, the fitness (i.e. the degree of suitability) of each component in the current solution is evaluated under an evaluation function. Then, the fitness measure is used probabilistically to select components to be eliminated in the Elimination-I step. Components with high fitness have a lower probability of being eliminated. Furthermore, to escape local minima in the solution space, capabilities for uphill moves must be incorporated. This is carried out in the Elimination-II step by probabilistically eliminating even some superior components of the solution in a totally random manner.

The resulting partial solutions are then fed into the Reconstruction step, which implements application specific heuristics to derive a new and complete solution from partial solutions. Throughout these iterations, the best solution is retained and finally returned as the final solution. This algorithm uses a greedy search strategy to achieve improvement through iterative perturbation and reconstruction.

___________________________________________________________________

CHEE ( )

{

   t = 0;

   Create an initial solution $S(0)$ with an associate cost $C(0)$;

   $C_{best}$= $C(0)$;

   While (stopping conditions not reached) {

      /* Decompose the solution into its component (i.e. shift patterns of individual nurses) */

      $S(t)$={$s_1, s_2,..., s_n$};

      /* The Evaluation step

      Use an evaluation function to assign each component a score;

      /* The Elimination-I step

      Eliminate some well-arranged components from $S(t)$;

      Obtain an incomplete solution $S'(t)$;

      /* The Elimination-II step

Randomly eliminate some components from $S'(t)$;

/* The Reconstruction step

Add new components into $S'(t)$ to make it complete;

$S(t) = S'(t)$;

If ($C(t)$ is better than $C_{best}$) $C_{best} = C(t)$;

t = t+1;

}

Return the best solution with the cost $C_{best}$;

}

---

Figure 1: The Pseudo Code of the Basic Algorithm

In summary, our methodology differs from some other local search methods such as simulated annealing (Kirkpatrick et al. 1983) and tabu search (Glover 1989) in the way that it does not follow one trajectory in the search space. By systematically eliminating components of a solution and then replenishing with new components, this algorithm essentially employs a longer sequence of moves between iterations, thus permitting more complex and more distant changes between successive solutions. This feature means that our method has the ability to jump quite easily out of local minima. Furthermore, unlike population-based evolutionary algorithms which need to maintain a number of solutions as parents for offspring propagation in each generation, this method operates on a single solution at a time. Thus, it should be able to eliminate the extra CPU-time needed to maintain a set of solutions.

# 3    A Component Based Heuristic Procedure with Evolutionary Elimination for Nurse Rostering

The basic idea behind the method is to determine, for each current schedule, the fitness of shift patterns assigned to individual nurses. The process keeps the shift patterns of some nurses that are well chosen (having high fitness values) in the current schedule and tries to replace the shift patterns of other nurses that have low fitness values. To enable the algorithm to execute iteratively, at each iteration, a randomly-produced threshold (in the range [0, 1]) is generated, and all shift patterns whose fitness values exceed the threshold are labelled as "good patterns" and survive in the current schedule. The remaining shift patterns are labelled as "bad patterns" and do not survive (become extinct). The fitness value therefore corresponds to the survival chance of a shift pattern assigned to a specific nurse. The "bad" shift patterns are removed from the current schedule and the corresponding nurses are released, waiting for their new assignments by a constructive heuristic. Following this, the above steps are iterated. Thus the global scheduling procedure is based on iterative improvement, while an iterative constructive process is performed within.

## 3.1 Initialization

In this step, an initial solution is generated to serve as a seed for its iterative improvement. It is well known that for most meta-heuristic algorithms, the initialization strategy can have a significant influence on performance. Thus, normally, a significant effort will be made to generate a starting point that is as good as possible. For nurse rostering, there are a number of heuristic techniques that can be applied to produce good starting solutions.

For our methodology, due to the fact that the replacement rate in its first iteration is relatively high, the performance is generally independent of the quality of the initial solution. However, if the seed is already a relatively good solution, the overall computation time will decrease. Since the major purpose of this paper is to demonstrate the performance and general applicability of the proposed methodology, we deliberately generate an extremely poor initial solution by randomly assigning a shift pattern to each nurse. The steps described in section 3.2 to 3.5 are executed in sequence in a loop until a stopping condition (i.e. solution quality or the maximum number of iteration) is reached.

## 3.2 Evaluation

In this step, the fitness of individual nurses' assignments, based on complete schedules, is evaluated. The evaluation function should be normalized and hence can be formulated as

$$F(E_i) = \sum_{k=1}^{2} w_k f_k(E_i), \quad \forall i \in \{1, ..., n\}, \tag{4}$$

subject to

$$\sum_{k=1}^{2} w_k = 1. \tag{5}$$

where $E_i$ are the shift pattern assigned to the i-th nurse, n is the number of nurses, $f_1(E_i)$ and $f_2(E_i)$ is the contribution of $E_i$ towards the preference and the feasibility aspect of the solution respectively.

$f_1(E_i)$ evaluates the shift pattern assigned to a nurse in terms of the degree to which it satisfies the soft constraints (i.e. this nurse's preference on his/her assigned shift pattern). It can be formulated as

$$f_1(E_i) = \frac{p_{max} - p_{ij}}{p_{max} - p_{min}}, \quad \forall i \in \{1, ..., n\} , \qquad (6)$$

where $p_{ij}$ is the preference cost of nurse i working shift pattern j and $p_{max}$ and $p_{min}$ are the maximum and minimum cost values among the shift patterns of all nurses on the current schedule, respectively.

$f_2(E_i)$ evaluates how far the shift pattern assigned to a nurse satisfies the hard constraints (i.e. coverage requirement and grade demands). This can be formulated as

$$f_2(E_i) = \frac{c_{ij} - c_{min}}{c_{max} - c_{min}}, \quad \forall i \in \{1, ..., n\} , \qquad (7)$$

where $c_{ij}$ is the coverage contribution of nurse i working shift pattern j and $c_{max}$ and $c_{min}$ are the maximum and minimum coverage contribution values among the shift patterns of all nurses on the current schedule, respectively.

In a current schedule, the coverage contribution of each nurse's shift pattern is its contribution to the cover of all three grades, which can be calculated as the sum of grade one, two and three covered shifts that would become uncovered if the nurse does not work on this shift pattern. Therefore, we formulate $c_{ij}$ as

$$c_{ij} = \sum_{s=1}^{3} q_{is} (\sum_{k=1}^{14} a_{jk} d_{ks}) , \qquad (8)$$

where $q_{is} = 1$ if nurse i is of grade s or higher, 0 otherwise;

$a_{jk} = 1$ if shift pattern j covers period k, 0 otherwise;

$d_{ks} = 1$ if there is a shortage of nurses during period k of grade s (i.e. the coverage value without considering shift pattern j is smaller than demand $R_{ks}$), 0 otherwise.

## 3.3  Elimination-I

This step is to determine whether the i-th nurses' assignment (denoted as $E_i$, $\forall i \in \{1,...,n\}$) should be retained for the next iteration or whether it should be eliminated and the nurse placed in the queue waiting for the next rescheduling. This is done by comparing his/her assignment fitness $F(E_i)$ to a random number $r_s$ generated for each iteration in the range $[0, 1]$. If $F(E_i) \leq r_s$, then $E_i$ will be removed from the current schedule; otherwise $E_i$ will survive in its present position. The days and nights that the nurses' shift pattern covers are then released and updated for the next Reconstruction step (see below). By using this step, an assignment $E_i$ with a larger fitness value $F(E_i)$ has a proportionally higher probability of survival in the current schedule. This mechanism performs in a similar way to roulette wheel selection in genetic algorithms.

## 3.4  Elimination-II

Following the Elimination-I step, the shift pattern of each remaining nurse still has a chance to be eliminated from the partial schedule at a given rate of $r_m$. The days and nights that an eliminated shift pattern covers are then released for the next Reconstruction step. As usual for mutation operators, compared with the elimination rate in the Elimination-I step, the rate here should be relatively smaller to facilitate convergence. Otherwise, there will be no bias in the sampling, leading to a random restart type algorithm. From a series of experiments we found that $r_m \leq 5.0\%$ yields good results and hence is the value adopted by us for our experiments. This process is analogous to the mutation operator in a genetic algorithm. Note that our method uses its Elimination-II step to eliminate some fitter components and thus generate a new diversified solution indirectly.

## 3.5  Reconstruction

The Reconstruction step takes a partial schedule as the input, and produces a complete schedule as the output. Since the new schedule is based on iterative improvement from the previous schedule, all shift assignments in the partial schedule should remain unchanged. Therefore, the Reconstruction task is reduced to assigning shift patterns to all unscheduled nurses to complete a partial solution.

Based on the domain knowledge of nurse rostering, there are many rules that can be used to build schedules. For example, Aickelin and Dowsland (2003) introduce three building rules: a 'Cover' rule, a 'Contribution' rule and a 'Combined' rule. Since the last two rules are quite similar, in this paper we only apply the 'Cover rule and the 'Combined' rule to fulfil the Reconstruction task.

The 'Cover' rule is designed to achieve the feasibility of the schedule by assigning each unscheduled nurse the shift pattern that covers the largest number of uncovered shifts. For instance, assume that a shift pattern covers Monday to Friday night shifts. Further assume that the current requirements for the night shifts from Monday to Sunday are as follows: (-4, 0, +1, -3, -1, -2, 0), where negative symbol means undercover and positive means over-cover. The given shift pattern hence has a cover value of 3 as it covers the night shifts of Monday, Thursday and Friday. Note that for nurses of grade s, this rule only counts the shifts requiring grade s nurses as long as there is a single uncovered shift for this grade. If all shifts of grade s are covered, shifts of grade (s-1) are counted. This operation is necessary as otherwise higher graded nurses might fill lower graded demand first, leaving the higher graded demand unmet.

The 'Combined' rule is designed to achieve a balance between solution quality and feasibility by going through the entire set of feasible shift patterns for a nurse and assigning each one a score. The one with the highest (i.e. best) score is chosen. If there is more than one shift pattern with the best score, the first such shift pattern is chosen. The score of a shift pattern is calculated as the weighted sum of the nurse's preference cost $p_{ij}$ for that particular shift pattern and its contribution to the cover of all three grades. The latter is measured as a weighted sum of grade one, two and three uncovered shifts that would be covered if the nurse worked this shift pattern, i.e. the reduction in shortfall. More precisely and using the same notation as before, the score $S_{ij}$ of shift pattern j for nurse i is calculated as

$$S_{ij} = w_p(100 - p_{ij}) + \sum_{s=1}^{3} w_s q_{is} (\sum_{k=1}^{14} a_{jk} e_{ks}),$$ (9)

where $w_p$ is the weight of the nurse's preference cost $p_{ij}$ for the shift pattern and $w_s$ is the weight of covering an uncovered shift of grade s. $q_{is}$ is 1 if nurse i is of grade s or higher, 0 otherwise. $a_{jk}$ is 1 if shift pattern j covers day k, 0 otherwise. $e_{ks}$ is the number of nurses needed to at least satisfy the demand $R_{ks}$ if there are still nurses in shortage during period k of grade s, 0 otherwise. $(100-p_{ij})$ must be used in the score, as higher $p_{ij}$ values are worse and the maximum for $p_{ij}$ is 100.

Using the above two rules at the rates of $p_1$ and $p_2$ respectively, the Reconstruction step assigns shift patterns to all unscheduled nurses until the partial solution is complete. In addition, to avoid stagnation at local optima, randomness needs to be introduced into the Reconstruction steps. This is achieved by allowing each unscheduled nurse to have an additional small rate $p_3$ to be scheduled by a randomly-selected shift pattern. Note that the sum of $p_1$, $p_2$ and $p_3$ should be 1. Also note that because we solve the problem without relying on any prior knowledge about which nurses should be scheduled earlier and which nurses later, the indexing order of nurses given in the original data set will be applied throughout the Reconstruction step.

After a partial solution is repaired, the fitness of this complete solution has to be calculated. Unfortunately, due to the highly-constrained nature of the problem, feasibility cannot be guaranteed. Hence, the following penalty function approach is used to evaluate the solutions obtained

$$\text{Min} \ \sum_{i=1}^{n} \sum_{j=1}^{m} p_{ij} x_{ij} + w_{demand} \sum_{k=1}^{14} \sum_{s=1}^{g} \max \left[ R_{ks} - \sum_{i=1}^{n} \sum_{j=1}^{m} q_{is} a_{jk} x_{ij} ; 0 \right], \quad (10)$$

where constant $w_{demand}$ is the penalty per uncovered shifts in the solution, and a "max" function is used due to the penalization of undercovering.

# 4  Computational Results

This section describes the computational experiments used to test our proposed algorithm. For all experiments, 52 real data sets (as provided by the hospital) are available. Each data set consists of one week's requirements (i.e. 14 time periods) for all shift and grade combinations and a list of nurses available together with their preference costs $p_{ij}$ and qualifications. Typically, there will be between 20 and 30 nurses per ward, 3 grade-bands and 411 different shift patterns. They are moderately sized problems compared to other problems reported in the literature (Burke et al. 2004). The data was collected from three wards over a period of several months and covers a range of scheduling situations, e.g. some data instances have very few feasible solutions whilst others have multiple optima. A zip file containing all these 52 instances is available to download at http://www.cs.nott.ac.uk/~jpl/Nurse_Data/NurseData.zip.

## 4.1  Algorithm Details

Table 1 lists detailed computational results of various approaches over 52 instances. The results listed in Table 1 are based on the best result out of 20 runs for each method with different random seeds. The second last row (headed 'Av.') contains the mean values of all columns, and the last row (headed '%') shows the relative percentage deviation values of the above mean values to the optimal solution values. When computing the mean, a censored cost value of 255 has been used if an algorithm fails to find a feasible solution (denoted as N/A). The following notations are employed in the table:

- IP: optimal or best-known solutions found by XPRESS MP, a commercial integer programming solver (Dowsland and Thompson 2000);

- GA-1: basic genetic algorithm reported in (Aickelin and White 2004);

- GA-2: adaptive GA, which is the same as GA-1, but it also tries to self-learn good parameters during the runtime starting from the values given below (Aickelin and White 2004);

- GA-3: multi-population genetic algorithm, which is the same as GA-2, but also features competing sub-populations (Aickelin and White 2004);

- GA-4: hill-climbing genetic algorithm, which is the same as GA-3, but it also includes a local search in the form of a hill-climber around the current best solution (Aickelin and White 2004);

- GA-5: indirect genetic algorithm, which maps the constraint solution space into an unconstrained space, then searches within that new space and eventually translates solutions back into the original space (Aickelin and Dowsland 2003). Up to four different rules and a hill-climber are used in this algorithm;

- EDA: estimation of distribution algorithm reported in (Aickelin and Li 2007);

- LCS: learning classifier system reported in (Li and Aickelin 2004);

- Con-heu: our method without the two steps of elimination;

- CHEE: our full Component based Heuristic method with both evolutionary perturbation steps;

  - Best: best result out of 20 runs of CHEE;

  - Mean: average result of 20 runs of CHEE;

  - Inf: number of runs terminating with the best solution being infeasible;

  - #: number of runs terminating with the best solution being optimal;

  - ≤3: number of runs terminating with the best solution being within three cost units of the optimum. The value of three units was chosen as it corresponds to the penalty cost of violating the least important level of requests in the original formulation. Thus, these solutions are still acceptable to the hospital.

Table 1: Comparison of Results by Various Approaches over 52 Instances

| Set | IP | GA-1 | GA-2 | GA-3 | GA-4 | GA-5 | EDA | LCS | Con-heu | CHEE (20 runs) | | | | |
|-----|-----|------|------|------|------|------|-----|-----|---------|------|------|-----|-----|-----|
| | | | | | | | | | | Best | Mean | Inf | # | ≤3 |
| 01 | 8 | 9 | 9 | 8 | 8 | 8 | 8 | 9 | 31 | 8 | 8.0 | 0 | 20 | 20 |
| 02 | 49 | 57 | 57 | 50 | 50 | 51 | 56 | 60 | 100 | 50 | 54.2 | 0 | 0 | 8 |
| 03 | 50 | 51 | 51 | 50 | 50 | 51 | 50 | 68 | 94 | 50 | 51.0 | 0 | 14 | 18 |
| 04 | 17 | 17 | 17 | 17 | 17 | 17 | 17 | 17 | 20 | 17 | 17.0 | 0 | 20 | 20 |
| 05 | 11 | 12 | 11 | 11 | 11 | 11 | 11 | 15 | 22 | 11 | 11.0 | 0 | 20 | 20 |
| 06 | 2 | 7 | 7 | 2 | 2 | 2 | 2 | 2 | 20 | 2 | 2.1 | 0 | 19 | 20 |
| 07 | 11 | N/A | N/A | 11 | 13 | 12 | 14 | 31 | 45 | 11 | 11.1 | 0 | 9 | 20 |
| 08 | 14 | 18 | 18 | 15 | 14 | 15 | 15 | 43 | 41 | 14 | 15.3 | 0 | 12 | 18 |
| 09 | 3 | N/A | N/A | 3 | 3 | 4 | 14 | 17 | N/A | 3 | 4.0 | 0 | 13 | 19 |
| 10 | 2 | 6 | 6 | 4 | 2 | 3 | 2 | 5 | 13 | 2 | 2.6 | 0 | 9 | 20 |
| 11 | 2 | 4 | 4 | 2 | 2 | 2 | 2 | 2 | N/A | 2 | 2.0 | 0 | 20 | 20 |
| 12 | 2 | 14 | 14 | 2 | 2 | 2 | 3 | 4 | N/A | 2 | 2.0 | 0 | 20 | 20 |
| 13 | 2 | 3 | 3 | 2 | 2 | 2 | 3 | 5 | 103 | 2 | 2.0 | 0 | 20 | 20 |
| 14 | 3 | 4 | 4 | 3 | 3 | 3 | 4 | 17 | 21 | 3 | 5.8 | 0 | 15 | 15 |
| 15 | 3 | 6 | 6 | 3 | 3 | 3 | 4 | 5 | 5 | 3 | 3.0 | 0 | 20 | 20 |
| 16 | 37 | 40 | 40 | 38 | 38 | 39 | 38 | 38 | 159 | 37 | 37.0 | 0 | 20 | 20 |
| 17 | 9 | 12 | 12 | 9 | 9 | 10 | 9 | 22 | N/A | 9 | 10.3 | 0 | 2 | 20 |
| 18 | 18 | 19 | 19 | 19 | 19 | 18 | 19 | 33 | 125 | 18 | 18.9 | 0 | 18 | 18 |
| 19 | 1 | 5 | 5 | 1 | 1 | 1 | 10 | 32 | N/A | 1 | 11.5 | 0 | 11 | 12 |
| 20 | 7 | 10 | 10 | 8 | 8 | 7 | 7 | 7 | 36 | 7 | 9.3 | 0 | 15 | 17 |
| 21 | 0 | 7 | 7 | 0 | 0 | 0 | 1 | 6 | 23 | 0 | 0.2 | 0 | 18 | 20 |
| 22 | 25 | 43 | 35 | 26 | 25 | 25 | 26 | 38 | 150 | 25 | 26.5 | 0 | 7 | 18 |
| 23 | 0 | 8 | 8 | 0 | 0 | 0 | 1 | 3 | N/A | 0 | 0.7 | 0 | 6 | 20 |
| 24 | 1 | 4 | 3 | 1 | 1 | 1 | 1 | 1 | N/A | 1 | 1.0 | 0 | 20 | 20 |
| 25 | 0 | 6 | 5 | 0 | 0 | 0 | 0 | 0 | 4 | 0 | 0.4 | 0 | 13 | 20 |
| 26 | 48 | N/A | N/A | 48 | 48 | 48 | 52 | 93 | 148 | 48 | 63.3 | 0 | 11 | 17 |
| 27 | 2 | 17 | 17 | 2 | 2 | 4 | 28 | 19 | N/A | 2 | 18.2 | 0 | 9 | 9 |

| 28 | 63 | 66 | 66 | 63 | 63 | 64 | 65 | 67 | N/A | 63 | 63.2 | 0 | 16 | 20 |
|---|---|---|---|---|---|---|---|---|---|---|---|---|---|---|
| 29 | 15 | 20 | 20 | 141 | 17 | 15 | 109 | 56 | N/A | 15 | 104.1 | 0 | 2 | 2 |
| 30 | 35 | 44 | 44 | 42 | 35 | 38 | 38 | 41 | 97 | 35 | 80.5 | 2 | 4 | 5 |
| 31 | 62 | N/A | 284 | 166 | 95 | 65 | 159 | 123 | N/A | 66 | 116.3 | 0 | 0 | 0 |
| 32 | 40 | 51 | 51 | 99 | 41 | 42 | 43 | 42 | N/A | 40 | 42.6 | 0 | 11 | 17 |
| 33 | 10 | N/A | N/A | 10 | 12 | 12 | 11 | 15 | N/A | 10 | 11.1 | 0 | 4 | 20 |
| 34 | 38 | 42 | 42 | 48 | 40 | 39 | 41 | 70 | N/A | 38 | 57.9 | 1 | 1 | 11 |
| 35 | 35 | 36 | 36 | 35 | 35 | 36 | 46 | 64 | N/A | 35 | 41.2 | 0 | 1 | 5 |
| 36 | 32 | N/A | 36 | 41 | 33 | 32 | 45 | 54 | 198 | 32 | 36.0 | 0 | 4 | 14 |
| 37 | 5 | 8 | 8 | 5 | 5 | 5 | 7 | 12 | 62 | 5 | 5.5 | 0 | 11 | 20 |
| 38 | 13 | N/A | N/A | 14 | 16 | 15 | 25 | 30 | 121 | 13 | 24.6 | 0 | 2 | 9 |
| 39 | 5 | 9 | 8 | 5 | 5 | 5 | 8 | 13 | 118 | 5 | 5.2 | 0 | 18 | 20 |
| 40 | 7 | 14 | 10 | 8 | 8 | 7 | 8 | 15 | 26 | 7 | 8.7 | 0 | 17 | 18 |
| 41 | 54 | N/A | 65 | 54 | 54 | 55 | 55 | 57 | 121 | 54 | 54.1 | 0 | 19 | 20 |
| 42 | 38 | 41 | 41 | 38 | 38 | 39 | 41 | 80 | 51 | 38 | 41.6 | 0 | 4 | 18 |
| 43 | 22 | 24 | 24 | 39 | 24 | 23 | 23 | 58 | N/A | 22 | 23.0 | 0 | 18 | 18 |
| 44 | 19 | 36 | 36 | 19 | 48 | 25 | 24 | 34 | N/A | 19 | 28.3 | 0 | 3 | 4 |
| 45 | 3 | N/A | 9 | 3 | 3 | 3 | 6 | 15 | 111 | 3 | 9.1 | 0 | 12 | 18 |
| 46 | 3 | 17 | 10 | 3 | 6 | 6 | 7 | 28 | N/A | 3 | 29.9 | 2 | 2 | 15 |
| 47 | 3 | N/A | 5 | 4 | 3 | 3 | 3 | 3 | N/A | 3 | 3.0 | 0 | 20 | 20 |
| 48 | 4 | 9 | 9 | 6 | 4 | 4 | 5 | 18 | N/A | 4 | 5.9 | 0 | 6 | 17 |
| 49 | 27 | 36 | 36 | 30 | 29 | 30 | 30 | 37 | N/A | 27 | 29 | 0 | 5 | 20 |
| 50 | 107 | N/A | N/A | 211 | 110 | 110 | 109 | 110 | N/A | 107 | 108.1 | 0 | 11 | 20 |
| 51 | 74 | N/A | N/A | N/A | 75 | 74 | 171 | 125 | N/A | 96 | 167.9 | 0 | 0 | 0 |
| 52 | 58 | N/A | N/A | N/A | 75 | 58 | 67 | 85 | N/A | 58 | 67.9 | 0 | 4 | 4 |
| Av. | 21.1 | 79.8 | 65.0 | 37.1 | 23.2 | 22.0 | 29.7 | 35.5 | 157.4 | 21.7 | 29.9 | 0.1 | 11.1 | 15.85 |
| % | 0 | 278 | 208 | 76 | 10 | 4 | 41 | 68 | 646 | 2.7 | 13.2 | | | |

For all data instances, we used the following set of fixed parameters in our experiments:

- Stopping criterion: a maximum iteration of 50,000, or an optimal/best-known solution has been found;

- Rate of Elimination-II in Section 3.4: $r_m$ =0.05;

- Rates of Reconstruction in Section 3.5: $p_1$ =0.80, $p_2$ =0.18, $p_3$ =0.02;

- Weight set in formula (9): $w_p$ =1, $w_1$ =8, $w_2$ =2 and $w_3$ =1;

- Penalty weight in fitness function (10): $w_{demand}$ =200.

Note that some parameter values (i.e. the maximum number of iterations, $r_m$, $p_1$, $p_2$ and $p_3$) are based on our experience and intuition and thus we cannot prove they are the best for each instance. The rest of the values (i.e. $w_p$, $w_1$, $w_2$, $w_3$ and $w_{demand}$ ) are the same as those used in previous papers for solving the same 52 instances, and we are continuing to use them for consistency.

Our method was coded in Java 2, and all experiments were undertaken on a Pentium 4 2.1GHz machine under Windows XP. To test the robustness of the proposed algorithm, each data instance was run twenty times by fixing the above parameters and varying the pseudo random number seed at the beginning. The execution time per run and per data instance varies from several milliseconds to 20 seconds depending on the difficulty of the individual data instance. Table 2 lists the average runtimes of various approaches over the same 52 instances: the first six (i.e. IP, GA-1, GA-2, GA-3, GA-4 and GA-5) were run on a different Pentium III PC, while the following two (i.e. EDA and LCS) on a similar Pentium 4 2.0GHz PC. Obviously, the IP is much slower than any of the above meta-heuristics. Among these meta-heuristic methods, our algorithm takes no more time although an accurate comparison in terms of runtime is difficult due to the different environments (i.e. machines, compilers and programming languages) in use. For example, the genetic algorithms are coded in C and the EDA is coded in C++. The comparison in terms of the number of evaluations is also difficult because the other algorithms evaluate each candidate solution as a whole, while our algorithm evaluates partial solutions as well.

Table 2: Comparison of the Average Runtime of Various Approaches

|  | IP | GA-1 | GA-2 | GA-3 | GA-4 | GA-5 | EDA | LCS | CHEE |
|---|---|---|---|---|---|---|---|---|---|
| Time (sec) | >24hours | 19 | 23 | 13 | 15 | 12 | 22 | 42 | 12 |

Table 3 lists the average results of 20 runs of CHEE under different parameter settings. Its first five columns contain the results after different maximum number of iterations, namely 10,000-20,000-30,000-50,000-100,000. Its last five columns contain the results of evaluating individual parts of CHEE, namely "Elimination-I ($r_s$=0.5) + Elimination-II + Con-heu", "Elimination-I + Con-heu", "Elimination-II + Con-heu", "Elimination I + Elimination II + Con-with-Cover-only" and "Elimination I + Elimination II + Con-with-Combined-only".

Table 3: Results of CHEE under Different Parameter Settings

| Set | Max number of iterations | | | | | Evaluation on individual parts (after $5\times10^4$ iterations) | | | | |
|---|---|---|---|---|---|---|---|---|---|---|
| | $10^4$ | $2\times10^4$ | $3\times10^4$ | $5\times10^4$ | $10^5$ | Pert-I ($r_s$=0.5) | Pert-I only | Pert-II only | Con-cover rule only | Con-combined rule only |
| 1 | 8.2 | 8.0 | 8.0 | 8.0 | 8.0 | 9.3 | 9.6 | 8.0 | 8.9 | 8.0 |
| 2 | 57.4 | 55.7 | 54.4 | 54.2 | 52.8 | 55.4 | 61.9 | 56.6 | 62.1 | 372.3 |
| 3 | 53.8 | 52.7 | 50.9 | 51.0 | 50.1 | 58.7 | 66.3 | 50.0 | 77.1 | 434.8 |
| 4 | 17.0 | 17.0 | 17.0 | 17.0 | 17.0 | 17.2 | 17.5 | 17.0 | 17.1 | 17.0 |
| 5 | 11.4 | 11.0 | 11.0 | 11.0 | 11.0 | 14.5 | 16.6 | 11.0 | 26.1 | 11.0 |
| 6 | 2.2 | 2.1 | 2.1 | 2.1 | 2.1 | 8.7 | 3.35 | 22.4 | 101.1 | 2.0 |
| 7 | 72.7 | 34.5 | 30.6 | 11.1 | 11.5 | 34.4 | 103.8 | 70.6 | 27.4 | 191.6 |
| 8 | 19.6 | 16.9 | 17.9 | 15.3 | 14.6 | 27.0 | 28.3 | 26.2 | 46.8 | 14.7 |
| 9 | 6.4 | 6.9 | 3.4 | 4.0 | 3.1 | 25.4 | 11.4 | 24.2 | 59 | 10.8 |
| 10 | 4.2 | 2.8 | 2.7 | 2.55 | 3.0 | 4.3 | 5.4 | 18.1 | 10.5 | 2.7 |
| 11 | 2.1 | 2.1 | 2.0 | 2.0 | 2.0 | 3.2 | 5.3 | 2.0 | 6.5 | 2.0 |
| 12 | 2.4 | 12.1 | 2.0 | 2.0 | 2.0 | 8.1 | 47.5 | 7.1 | 2.4 | 2.0 |
| 13 | 34.9 | 3.1 | 2.0 | 2.0 | 2.0 | 5.5 | 4.6 | 2.15 | 53.7 | 2.0 |
| 14 | 13.7 | 12.9 | 12.0 | 5.75 | 4.2 | 30.3 | 20.8 | 100.0 | 161.9 | 3.5 |
| 15 | 10.7 | 3.1 | 3.0 | 3.0 | 3.0 | 4.9 | 4.8 | 3.0 | 21.9 | 3.0 |
| 16 | 37.4 | 37.7 | 37.0 | 37.0 | 37.0 | 63.0 | 38.5 | 107.5 | 130.6 | 425.0 |
| 17 | 21.4 | 10.1 | 10.0 | 10.3 | 9.9 | 22.7 | 41.1 | 47.2 | 106.4 | 32.9 |
| 18 | 48.6 | 20.7 | 18.1 | 18.9 | 18.0 | 108.7 | 27.45 | 210.7 | 61.1 | 60.4 |

| 19 | 16.2 | 14.4 | 15.9 | 11.5 | 7.2 | 29.7 | 64.6 | 48.1 | 57.5 | 100.6 |
|----|------|------|------|------|-----|------|------|------|------|-------|
| 20 | 20.4 | 13.5 | 9.8 | 9.3 | 10.8 | 14.0 | 15.4 | 46.2 | 11.6 | 205.0 |
| 21 | 20.4 | 20.1 | 20 | 0.2 | 0.0 | 1.3 | 29.6 | 15.0 | 21.5 | 10.1 |
| 22 | 29.6 | 26.9 | 27.1 | 26.5 | 25.2 | 34.4 | 52.6 | 31.5 | 31.8 | 26.0 |
| 23 | 10.0 | 0.5 | 0.8 | 0.7 | 0.5 | 3.2 | 4.5 | 40.3 | 220.8 | 50.2 |
| 24 | 1.0 | 1.0 | 1.0 | 1.0 | 1.0 | 2.1 | 1.1 | 1.0 | 26.7 | 1.0 |
| 25 | 0.7 | 0.4 | 0.6 | 0.4 | 0.1 | 1.0 | 2.2 | 0.9 | 1.1 | 0.4 |
| 26 | 198.8 | 148.5 | 138.4 | 63.3 | 48.2 | 183.4 | 207.8 | 158.5 | 153.9 | 229.1 |
| 27 | 16.6 | 18.9 | 6.7 | 18.2 | 7.6 | 83.9 | 45.8 | 80.9 | 41.9 | 71.8 |
| 28 | 63.7 | 63.3 | 63.6 | 63.2 | 63.1 | 65.1 | 70.8 | 63.0 | 68.7 | 63.0 |
| 29 | 114.6 | 114.4 | 114.1 | 104.1 | 114.0 | 152.5 | 111.7 | 148.4 | 207.6 | 215.0 |
| 30 | 138.7 | 125.2 | 146.5 | 80.5 | 43.3 | 160.0 | 116.6 | 164.2 | 212.7 | 307.3 |
| 31 | 135.3 | 123.9 | 123.0 | 116.3 | 105.3 | 221.7 | 151.3 | 226.6 | 158.4 | 468.5 |
| 32 | 89.9 | 52.1 | 44.8 | 42.6 | 44.6 | 165.2 | 67.05 | 98.7 | 323.9 | 410.2 |
| 33 | 49.3 | 20.9 | 12.0 | 11.1 | 10.8 | 14.1 | 18.4 | 70.1 | 17.0 | 86.9 |
| 34 | 141.4 | 118.7 | 94.6 | 57.9 | 41.6 | 106.8 | 126.5 | 122.1 | 61.2 | 226.2 |
| 35 | 54.1 | 49.5 | 46.5 | 41.2 | 38.5 | 59.3 | 69.6 | 48.5 | 71.4 | 39.5 |
| 36 | 45.8 | 43.1 | 40.1 | 36.0 | 32.6 | 47.2 | 53.6 | 51.3 | 136.8 | 153.3 |
| 37 | 6.4 | 6.6 | 6.1 | 5.5 | 5.4 | 11.7 | 12.5 | 10.3 | 17.2 | 5.7 |
| 38 | 61.4 | 30.3 | 20.1 | 24.6 | 19.6 | 66.6 | 66.2 | 76.9 | 28.8 | 86.0 |
| 39 | 16.1 | 6.4 | 5.5 | 5.2 | 5.1 | 10.5 | 15.4 | 5.2 | 57.5 | 5.2 |
| 40 | 14.3 | 14.2 | 12.0 | 8.7 | 7.2 | 17.4 | 15.7 | 37.8 | 46.5 | 41.3 |
| 41 | 55.1 | 54.1 | 54.1 | 54.1 | 54.0 | 68.2 | 58.1 | 148.3 | 141.2 | 433.6 |
| 42 | 49.5 | 43.1 | 46.7 | 41.6 | 39.8 | 51.0 | 78.7 | 77.1 | 135 | 60.2 |
| 43 | 29.2 | 26.0 | 23.0 | 23.0 | 24.0 | 106.0 | 34.6 | 211.3 | 72.5 | 22.0 |
| 44 | 38.2 | 33.8 | 31.7 | 28.3 | 26.3 | 30.4 | 34.9 | 62.6 | 32.4 | 91.7 |
| 45 | 25.9 | 33.9 | 15.6 | 9.1 | 3.0 | 15.4 | 40.4 | 18.8 | 24.3 | 23.1 |
| 46 | 148.8 | 109.7 | 54.5 | 29.9 | 5.3 | 186.9 | 166.9 | 74.7 | 232.3 | 196.0 |
| 47 | 3.2 | 3.0 | 3.0 | 3.0 | 3.0 | 4.0 | 7.9 | 3.0 | 19.7 | 3.0 |
| 48 | 14.7 | 9.6 | 8.4 | 5.9 | 4.3 | 16.8 | 19.9 | 11.8 | 20.6 | 54.9 |

| 49 | 31.3 | 32.2 | 32.0 | 29.0 | 27.5 | 54.4 | 45.3 | 147.5 | 111.7 | 205.2 |
|---|---|---|---|---|---|---|---|---|---|---|
| 50 | 108.6 | 108.9 | 108.6 | 108.1 | 107.4 | 112.3 | 111.1 | 272.3 | 129.5 | 275.1 |
| 51 | 175.8 | 173.4 | 162.5 | 167.6 | 171.0 | 285.0 | 176.1 | 294.1 | 333.0 | 429.0 |
| 52 | 132.4 | 79.7 | 74.5 | 67.9 | 59.0 | 93.2 | 96.8 | 132.3 | 78.9 | 222.0 |
| Av. | 47.7 | 39.0 | 35.5 | 29.9 | 27.1 | 57.2 | 52.0 | 72.7 | 82.4 | 123.3 |

## 4.2  Analysis of Results

The results of all the approaches in Table 1 are obtained by using the same 52 benchmark test instances, with the bold figure representing the optimal solution found by a commercial software package.  Compared with the results of the mathematical programming approach which can take up to 24 hours runtime (shown in the 'IP' column), our results (shown in the 'Best' column) are only 2.7% more expensive on average but they are all achieved within 20 seconds.  Compared with the best results of various meta-heuristic approaches, in general the CHEE results are slightly better than those of the best-performing indirect genetic algorithm (with a relative percentage deviation value of 4%) and are much better than the others (with deviation values from 10% to 278%). A student's t-test (where "255" is used instead of "N/A") also suppose the observations: considering the best values for 52 instance, CHEE performs better tan GA-1, GA-2, GA-3, EDA and LCS within a confidence interval of 95%.

Since our proposed methodology uses a 'Cover' rule and a 'Combined' rule in its Reconstruction step for schedule repairing, it may be interesting to know if the good performance of our algorithm is mainly due to these two delicate building rules.  To clarify this, we performed an additional set of experiments by skipping the two elimination steps, i.e. only implementing the Reconstruction step to build a schedule from an empty solution.  This method does not yield a single feasible solution for 24 instances, as the 'Con-heu' column shows.  This underlines the difficulty of this problem, and most importantly it underlines the key roles played by the two elimination steps in our full methodology, as the Reconstruction step alone is not capable of solving the problem.

Table 2 shows the effect of the maximum number of iterations and the effect of each method with different parameter setting to the proposed CHEE. Clearly, the larger the maximum number of iterations, the better the solution quality we can obtain. Regarding the five methods within CHEE, "Elimination-I + Con-heu" performs best (with an average value of 52.0), "Elimination-I ($r_s$=0.5) + Elimination-II + Con-heu" performs second, "Elimination-II + Con-heu" performs third, "Elimination I + Elimination II + Con-with-Cover-only" performs fourth and "Elimination I + Elimination II + Con-with-Combined-only" performs worst. However, even the best performing "Elimination-I + Con-heu" method is much worse than a standard full CHEE method introduced in Section 3 (with an average value of 29.9). These results reveal that all the parts of CHEE are important to deliver high quality solutions and none of them could be freely removed. Figures 2 and 3 show the results of our method and the best indirect genetic algorithm graphically in more detail. The bars above the y-axis represent solution quality out of 20 runs: the black bars show the number of optimal solutions found (i.e. the value of '#' in Table 1), and the dotted bars represent the number of good feasible solutions which are within 3 cost units of their optimal solutions (i.e. the value of '≤3' in Table 1). The bars below the y-axis represent the number of times the algorithm failed to find a feasible solution in these 20 runs (i.e. the value of 'Inf' in Table 1). Hence, the smaller the area below the y-axis and the larger the area above, the better the algorithm's performance. Note that 'missing' bars mean that, over 20 runs, feasible solutions are obtained at least once, but none of them are optimal or of good quality (within 3 units of optimal values).

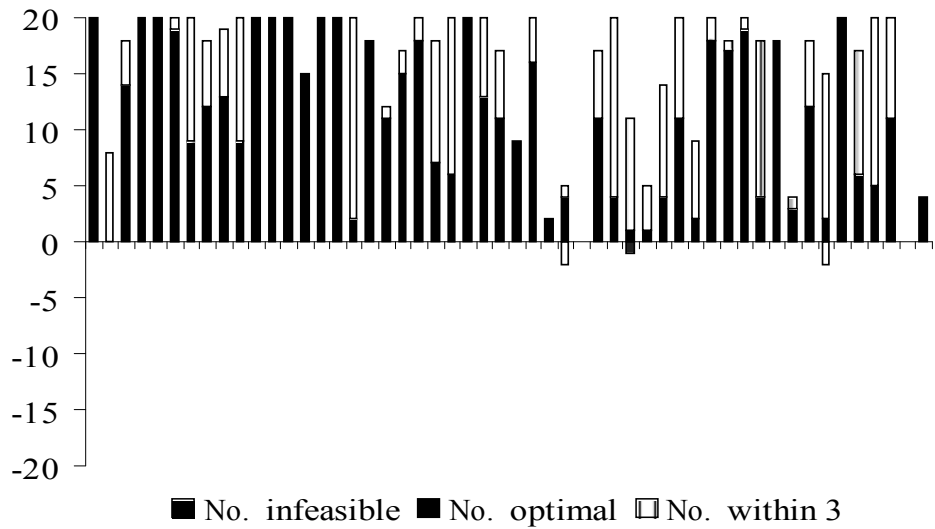

Figure 2: Results from CHEE

Figure 2 shows that 21 out of 52 data instances are solved well by CHEE (i.e. with all solutions being within 3 units of optimal values), 49 instances are solved optimally at least once, and overall there are 5 infeasible solutions for 3 instances. For the best indirect genetic algorithm (shown in figure 3), the results are slightly worse: 15 data instances are solved well, 28 are solved to optimality at least once, and overall there are 56 infeasible solutions for 6 data instances.

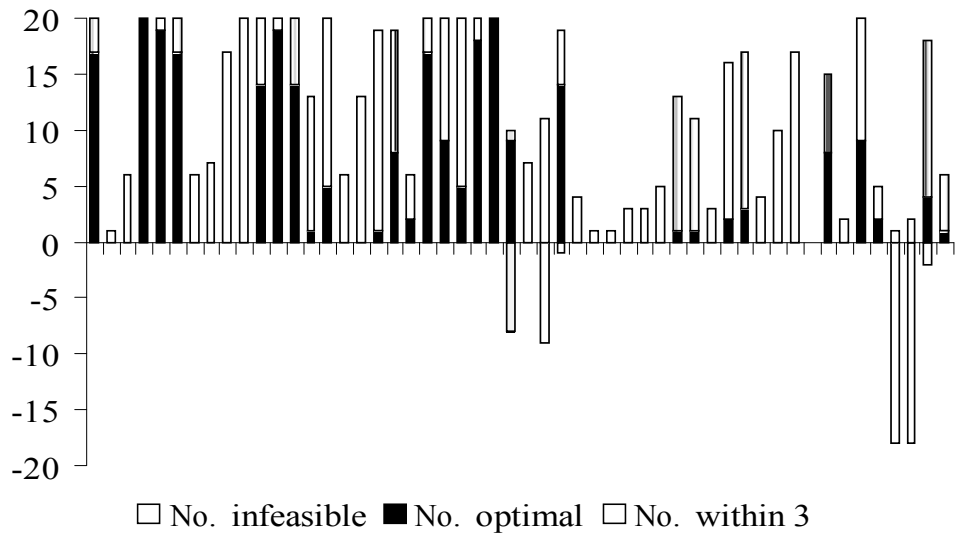

Figure 3: Results of the Best Indirect Genetic Algorithm (i.e. GA-5)

Figure 4 summarizes Table 1 in graphical format and provides an overall performance comparison between our proposed methodology and the other approaches. The best results for these instances are obtained by the IP software, and in general, our approach performs better than the previous best-performing meta-heuristic approach. The basic genetic algorithm (i.e. GA-1), the adaptive genetic algorithm (i.e. GA-2), the multi-population genetic algorithm (i.e. GA-3) and even the hill-climbing genetic algorithm (i.e. GA-4) which includes multiple populations and an elaborate local search are all significantly outperformed in terms of feasibility, best and average results.

The other three approaches (i.e. the GA-5, the EDA and the LCS) belong to the class of indirect approaches, in which a set of heuristic rules, including the 'Cover' rule and the 'Combined' rule used in our approach, is used for schedule building. Compared with the EDA and the LCS, our new approach performs much better in terms of the best and average results, and slightly worse in terms of feasibility. Compared with the GA-5 which performs best among all the heuristic algorithms, our approach performs better in all aspects of feasibility (99% vs. 95%), best results (21.7 versus 22.0) and average results (28.6 vs. 35.6). In addition, it is worth mentioning that the GA-5 uses the best possible order of the nurses (which, of course, has to be found) for the greedy heuristic to build a schedule, while our algorithm only uses a fixed indexing ordering given in the original data sets.

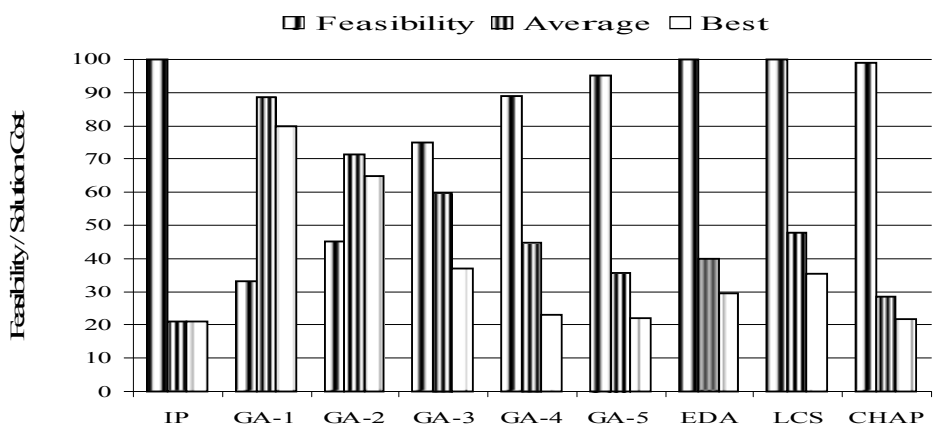

Figure 4: Summary Results of Various Search Algorithms

## 5 Conclusions

This paper presents a new approach to address the hospital personnel scheduling problem. The major idea behind this method is to decompose a solution into components, and then to implement two evolutionary-based elimination strategies on these components to make iterative improvements in each single schedule. In each iteration, an unfit portion of the solution is removed. Any partial solution is repaired by a refined greedy building process.

Taken as a whole, the proposed approach has a number of distinct advantages. Firstly, it is simple and easy to implement because it uses greedy algorithms and local heuristics. Secondly, due to its features of maintaining only a single solution at each iteration and eliminating inferior parts from this solution, it can quickly converge to local optima. Thirdly, the technique has the ability to jump out of local optima in an effective manner. Finally, this approach can be easily combined with other meta-heuristics to achieve its peak performance on solution quality if CPU-time is not the major concern. For example, tabu search can be used in the Reconstruction step to explore the neighbouring solutions in an aggressive way and avoid cycles by declaring attributes of visited solutions as tabu. In addition, simulated annealing could be used as the acceptance criteria for the resulting solutions after Reconstruction to accept not only improved solutions as in the current form, but also worse ones with a certain level of probability.

## Acknowledgements

The work was funded by the UK Government's major funding agency, the Engineering and Physical Sciences Research Council (EPSRC), under grants GR/R92899/02 and GR/S70197/1.

## References

Ahuja, R.K., T.L. Magnanti, J.B. Orlin. 1993. Network Flows: Theory, Algorithms, and Applications. Prentice Hall, NJ.

Aickelin, U., K. Dowsland. 2000. Exploiting problem structure in a genetic algorithm approach to a nurse rostering problem. Journal of Scheduling 3 139–153.

Aickelin, U., K. Dowsland. 2003. An indirect genetic algorithm for a nurse scheduling problem. Computers and Operations Research 31 761–778.

Aickelin, U., J. Li. 2007. An estimation of distribution algorithm for nurse scheduling. Annals of Operations Research 155 289–309.

Aickelin, U., P. White. 2007. Building better nurse scheduling algorithms. Annals of Operations Research 128 159–177.

Anzai, M., Y. Miura. 1987. Computer program for quick work scheduling of nursing staff. Medical Informatics 12 43–52.

Bard, J., H.W. Purnomo. 2006. Preference scheduling for nurses using column generation. European Journal of Operational Research 164 510–534.

Bard, J., H.W. Purnomo. 2007. A cyclic preference scheduling of nurses using a Lagrangian-based heuristic. Journal of Scheduling 10 5–23.

Beddoe, G., S. Petrovic. 2006. Selecting and weighting features using a genetic algorithm in a case-based reasoning approach to personnel rostering. (to appear) European Journal of Operational Research.

Beliën, J., E.L. Demeulemeester. 2006. Building cyclic master surgery schedules with leveled resulting bed occupancy. European Journal of Operational Research 176 1185-1204.

Blau, R. 1985. Multishift personnel scheduling with a microcomputer. Personnel Administrator 20 43–58.

Bradley, D., J. Martin..1990. Continuous personnel scheduling algorithms: a literature review. Journal of the Society for Health Systems 2 8–23.

Brusco, M.J., L.W. Jacobs. 1995. Cost analysis of alternative formulations for personnel scheduling in continuously operating organisations. European Journal of Operational Research 86 249–261.

Burke, E.K., P. Cowling, P. De Causmaecker, G. Vanden Berghe. 2001. A memetic approach to the nurse rostering problem. Applied Intelligence 15 199–214.

Burke, E.K., P. De Causmaecker, G. Vanden Berghe. 1999. A hybrid tabu search algorithm for the nurse rostering problem. B. McKay et al., eds. Simulated Evolution and Learning. Springer Springer Lecture Notes in Computer Science Volume 1585 187–194.

Burke, E.K., P. De Causmaecker, S. Petrovic, G. Vanden Berghe. 2004. Variable neighbourhood search for nurse rostering problems. M.G.C. Resende, J.P. De Sousa, eds. Metaheuristics: Computer Decision-Making (Combinatorial Optimization Book Series). Kluwer, Chapter 7. 153–172.

Burke, E.K., P. De Causmaecker, G. Vanden Berghe, H. Van Landeghem. 2004. The state of the art of nurse rostering. Journal of Scheduling 7 441–499.

Burke, E.K., G. Kendall, J. Newall, E Hart, P. Ross, S. Schulenburg. 2003. Hyper-heuristics: an emerging direction in modern search technology. F. Glover, G. Kochenberger, eds. Handbook of Meta-Heuristics. Kluwer, Chapter 16. 451–470.

Burke, E.K., G. Kendall, E. Soubeiga. 2003. A tabu-search hyperheuristic for timetabling and rostering. Journal of Heuristics 9 451–470.

Cheang, B., H. Li, A. Lim, B. Rodrigues. 2003. Nurse rostering problems – a bibliographic survey. European Journal of Operational Research 151 447–460.

Chen, J.G., T. Yeung. 1993. Hybrid expert system approach to nurse scheduling. Computers in Nursing 183–192.

Dowsland, K. 1998. Nurse scheduling with tabu search and strategic oscillation. European Journal of Operational Research 106 393–407.

Dowsland, K., J. Thompson. 2000. Nurse scheduling with knapsacks, networks and tabu search. Journal of the Operational Research Society 51 825–833.

Ernst, A.T., H. Jiang, M. Krishnamoorthy, B. Owens, D. Sier. 2004a. An annotated bibliography of personnel scheduling and rostering. Annals of Operations Research 127 21–144.

Ernst, A.T., H. Jiang, M. Krishnamoorthy, D. Sier. 2004b. Staff scheduling and rostering: a review of applications, methods and models. European Journal of Operational Research 153 3–27.

Glover, F. 1989. Tabu search – part I. ORSA Journal on Computing 1 190–206.

Kawanaka, H., K. Yamamoto, T. Yoshikawa, T. Shinigi, S. Tsuruoka. 2001. Genetic algorithm with the constraints for nurse scheduling problem. Proceedings of Congress on Evolutionary Computation. 1123-1130.

Kirkpatrick, S., C.D. Gelatt, M.P. Vecchi. 1983. Optimization by simulated annealing. Science 220 671–680.

Li, J., U. Aickelin. 2004. The application of Bayesian optimization and classifier systems in nurse scheduling. X. Yao, et al. eds. Proceedings of the 8th International Conference on Parallel Problem Solving from Nature (PPSN VIII). Springer Lecture Notes in Computer Science Volume 3242 581–590.

Lourenco, H.R., O.C. Martin, T. Stutzle. 2002. Iterated local search. F. Glover, G. Kochenberger, eds. Handbook of Metaheuristics. Kluwer, ISORMS 57.. 312–353.

Meyer auf'm Hofe, H. 2001. Solving rostering tasks as constraint optimization. E.K. Burke, W. Erben, eds. Practice and Theory of Automated Timetabling, 3rd International Conference. Springer Lecture Notes in Computer Science Volume 2079 191–212.

Miller, H.E., W. Pierskalla, G. Rath. 1976. Nurse scheduling using mathematical programming. Operations Research 24 857–870.

Mueller, C.W., J.C. McCloskey. 1990. Nurses' job satisfaction: a proposed measure. Nursing Research 39 113-117.

Özcan, E. 2005. Memetic algorithms for nurse rostering. Proceedings of the 20th International Symposium on Computer and Information Sciences. Springer Lecture Notes in Computer Science Volume 3733 482-492.

Ross, P. 2005. Hyper-heuristics. E.K. Burke, G. Kendall, eds. Search Methodologies: Introductory Tutorials in Optimization and Decision Support Methodologies. Chapter 16, Springer.

Schrimpf G., J. Schneider, H. Stamm-Wilbrandt, G. Dueck. 2000. Record breaking optimization results using the ruin and recreate principle. Journal of Computational Physics 159 139-171.

Silvestro, R., C. Silvestro. 2000. An evaluation of nurse rostering practices in national health service. Journal of Advanced Nursing 32 525-535.

Sitompul, D., S. Randhawa. 1990. Nurse scheduling models: a state-of-the-art review. Journal of the Society of Health Systems 2 62-72.

Tien, J.M., A. Kamiyama. 1982. On manpower scheduling algorithms. Society for Industrial and Applied Mathematics 24 275–287.

Warner, M. J. Prawda. 1972. A mathematical programming model for scheduling nursing personnel in a hospital. Management Science 19 411–422.